%% file: eccv24_cameraready.tex
\documentclass[runningheads]{llncs}

\usepackage{eccv}

\usepackage{eccvabbrv}

\usepackage{graphicx}
\usepackage{booktabs}

\usepackage[accsupp]{axessibility}  %

\usepackage[dvipsnames]{xcolor}

\definecolor{cvprblue}{rgb}{0.21,0.49,0.74}

\usepackage{mathtools,amsmath, bm, epstopdf, pifont, overpic, cases}
\usepackage{latexsym, amssymb, bbding, multirow, makecell,  diagbox, enumitem, caption}
\usepackage{overpic,array, enumitem, soul, algorithm, algpseudocode}

\DeclareMathOperator*{\argmin}{arg\,min}
\usepackage{anyfontsize}
\usepackage{rotating}

\usepackage{hyperref}

\usepackage{orcidlink}

\begin{document}

\title{MoVideo: Motion-Aware Video Generation with Diffusion Model} 

\titlerunning{MoVideo: Motion-Aware Video Generation with Diffusion Model}

\author{Jingyun Liang\inst{1} \and Yuchen Fan\inst{2} \and Kai Zhang\inst{4}\thanks{Corresponding author.} \and Radu Timofte\inst{3} \and Luc Van Gool\inst{1} \and Rakesh Ranjan\inst{2}}

\authorrunning{J.~Liang et al.}

\institute{Computer Vision Lab, ETH Zurich, Switzerland \and Meta Inc.
\and Computer Vision Lab, CAIDAS \& IFI, University of Würzburg, Germany
\and Nanjing University, Suzhou, China\\
\email{\{jinliang,vangool\}@vision.ee.ethz.ch} \email{radu.timofte@uni-wuerzburg.de}
\\
\email{\{ycfan, rakeshr\}@meta.com}  \email{kaizhang@nju.edu.cn}
\url{https://jingyunliang.github.io/MoVideo}
} 

\maketitle

\renewcommand\twocolumn[1][]{#1}%
\input{sec_eccv_cameraready/teaser}
\input{sec_eccv_cameraready/0_abstract}
\input{sec_eccv_cameraready/1_intro}

\input{sec_eccv_cameraready/2_relatedwork}
\input{sec_eccv_cameraready/3_method}

\input{sec_eccv_cameraready/4_experiment}
\input{sec_eccv_cameraready/5_conclusion}

\section*{Acknowledgements}
Jingyun Liang and Luc Van Gool were partially supported by the ETH Zurich Fund (OK), a Huawei Technologies Oy (Finland) project and the China Scholarship Council. Radu Timofte was partially supported by The Alexander von Humboldt Foundation. Yuchen Fan and Rakesh Ranjan from Meta did not receive above fundings.

\bibliographystyle{splncs04}
\bibliography{main}

\end{document}

%% file: sec_eccv_cameraready/teaser.tex
\begin{center}
\begin{overpic}[width=12.2cm]{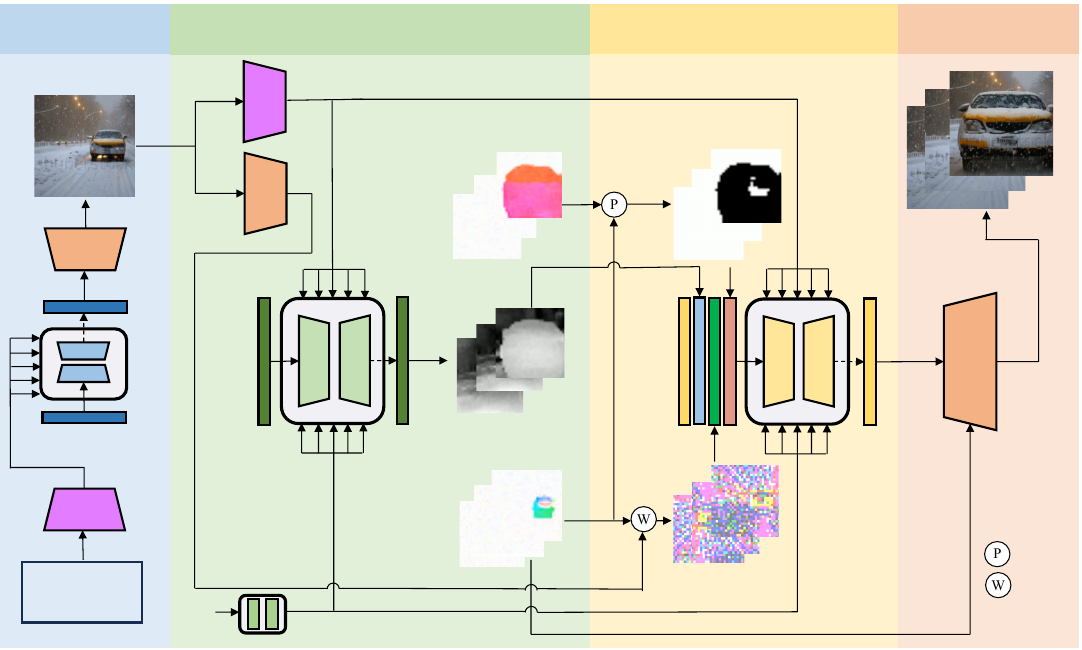}
\put(3,56.9){\color{black}{{\fontsize{5}{6}\selectfont \makecell{Key Frame\\ Generation}}}}
\put(25,56.9){\color{black}{{\tiny{\makecell{Video Depth and \\Optical Flow Generation}}}}}
\put(58.3,56.9){\color{black}{{\tiny{\makecell{Depth and Flow-based\\ Video Generation}}}}}
\put(84,56.9){\color{black}{{\tiny{\makecell{Flow-Augmented \\Video Decoding}}}}}
\put(4.5,52.2){\color{black}{{\scalebox{0.8}{\tiny{Key Frame}}}}}
\put(5,36.8){\color{black}{{\scalebox{0.8}{\tiny{\makecell{VQGAN\\Decoder}}}}}}
\put(5.1,12.6){\color{black}{{\scalebox{0.8}{\tiny{\makecell{Encoder\\(text)}}}}}}
\put(2.3,5.2){\color{black}{{\scalebox{0.8}{\tiny{\makecell{a car is moving\\on the snow}}}}}}
\put(5.3,1.2){\color{black}{{\scalebox{0.8}{\tiny{\makecell{Prompt}}}}}}
\put(22.8,53.5){\color{black}{{\scalebox{0.8}{\tiny\rotatebox{270}{\makecell{Encoder\\(image)}}}}}}
\put(22.8,45.7){\color{black}{{\scalebox{0.8}{\tiny\rotatebox{270}{\makecell{VQGAN\\(Encoder)}}}}}}
\put(17,3.1){\color{black}{{\scalebox{0.8}{\tiny{{$fps$}}}}}}
\put(38.5,45.2){\color{black}{{\scalebox{0.8}{\tiny\rotatebox{270}{\makecell{Video-to-Image\\Flow}}}}}}
\put(38.5,17){\color{black}{{\scalebox{0.8}{\tiny\rotatebox{270}{\makecell{Image-to-Video\\Flow}}}}}}
\put(43,20){\color{black}{{\scalebox{0.8}{\tiny{\makecell{Depth}}}}}}
\put(62,47.3){\color{black}{{\scalebox{0.8}{\tiny{\makecell{Occlusion Mask}}}}}}
\put(61.5,6.){\color{black}{{\scalebox{0.8}{\tiny{\makecell{Warped\\ Latent Video}}}}}}
\put(83.7,38.5){\color{black}{{\scalebox{0.8}{\tiny{\makecell{Generated\\Video}}}}}}
\put(88.7,32.2){\color{black}{{\scalebox{0.75}{\tiny\rotatebox{270}{\makecell{Flow-Augmented\\3D Decoder}}}}}}
\put(94,8.6){\color{black}{{\scalebox{0.8}{\tiny{\makecell{Predict}}}}}}
\put(94,5.6){\color{black}{{\scalebox{0.8}{\tiny{\makecell{Warp}}}}}}
\put(4.8,19.5){\color{black}{{\scalebox{0.8}{\tiny{${\mathcal{N}(0,1)}$}}}}}
\put(22,29.5){\color{black}{{\rotatebox{270}{\scalebox{0.8}{\tiny{${\mathcal{N}(0,1)}$}}}}}}
\put(60.9,29.5){\color{black}{{\rotatebox{270}{\scalebox{0.8}{\tiny{${\mathcal{N}(0,1)}$}}}}}}
\end{overpic}
\captionsetup{type=figure}
\caption{The schematic illustration of the proposed motion-aware video generation (MoVideo) framework. Given a text prompt, we first generate the key frame by a public available latent diffusion model. Then, we generate video depth and optical flows conditional on the image embedding (extracted by an open-sourced pretrained image-text bi-encoder model) and frames per second. Next, we add extra conditions, including depth, flow-based warped latent video and calculated occlusion mask, to generate the video in the latent space. Last, the video is decoded with flow-based alignment and feature refinement modules.}
\label{fig:teaser}
\end{center}

%% file: sec_eccv_cameraready/0_abstract.tex
\begin{abstract}
While recent years have witnessed great progress on using diffusion models for video generation, most of them are simple extensions of image generation frameworks, which fail to explicitly consider one of the key differences between videos and images, \ie, motion. In this paper, we propose a novel motion-aware video generation (MoVideo) framework that takes motion into consideration from two aspects: video depth and optical flow. The former regulates motion by per-frame object distances and spatial layouts, while the later describes motion by cross-frame correspondences that help in preserving fine details and improving temporal consistency. More specifically, given a key frame that exists or generated from text prompts, we first design a diffusion model with spatio-temporal modules to generate the video depth and the corresponding optical flows. Then, the video is generated in the latent space by another spatio-temporal diffusion model under the guidance of depth, optical flow-based warped latent video and the calculated occlusion mask. Lastly, we use optical flows again to align and refine different frames for better video decoding from the latent space to the pixel space. In experiments, MoVideo achieves state-of-the-art results in both text-to-video and image-to-video generation, showing promising prompt consistency, frame consistency and visual quality.
\end{abstract}

%% file: sec_eccv_cameraready/1_intro.tex
\section{Introduction}
\label{sec:intro}
In video generation, how to generate videos with natural consistent motions is one of the key challenges. In the era of deep learning, Generative
Adversarial Networks~\cite{goodfellow2014GAN,vondrick2016generating,saito2017temporal,tulyakov2018mocogan} (GANs) and autoregressive models~\cite{vaswani2017transformer,wu2021godiva,ge2022long,wu2022nuwa,hong2022cogvideo} have become two primary workhorses for video generation due to their great generative modelling abilities. However, GANs are hard to train and may suffer from model collapse, while autoregressive models represent a video as a sequence of tokens from a limited-sized dictionary, which might be insufficient to cover the general video domain. 

Recently, diffusion models~\cite{sohl2015deep,song2020denoising,song2020score} have attracted much attention due to its impressive performance on image generation~\cite{ho2022cascaded,ramesh2022hierarchical,saharia2022photorealistic,zhu2023denoising}. They design a forward diffusion process to gradually add noise to the image and a reverse process to gradually remove noise by a learned UNet model~\cite{ronneberger2015u}, under the assumption of the Markov chain with learned Gaussian transitions~\cite{ho2020denoising}. To apply diffusion models to video generation, one natural idea is to simply regard the video as a 3D extension of the 2D image and add an extra temporal dimension to the 2D UNet denoising network~\cite{ho2022video,singer2022make,esser2023structure,ho2022imagen,he2022latent,zhou2022magicvideo}. However, it might be challenging for the model to learn the video motions implicitly due to the lack of large-scale high-quality video datasets and the limited learning ability of 3D UNet models.

In this paper, we propose to explicitly model and utilize motion for video generation, incorporating depth and optical flow. Depth is employed to guide the per-frame spatial layouts, and a sequence of depth maps is used to capture the movements within the corresponding video. Optical flow, on the other hand, represents correspondences between different frames in the video and can be leveraged for frame alignment, preserving fine details and enhancing temporal consistency. More specifically, as shown in Fig.~\ref{fig:teaser}, given an existing key frame or an image generated by an image latent diffusion model, we first generate the depth and optical flow of the whole video, by utilizing a 3D diffusion model with spatio-temporal modelling blocks. After that, under the joint guidance of depth, optical flow-based warped latent video and the calculated occlusion mask, we generate the video in the latent space with another 3D spatio-temporal diffusion model. The final video is decoded to the pixel space with optical flow-based alignment and feature refinement.

Our contributions are summarized as follows:
\begin{itemize}[nosep,left=1em]
    \item [1)] To the best of our knowledge, we are the first to generate video depth and optical flows from texts or images. We found that a single static image holds clues about the movements of objects and backgrounds, showing great ability in generating video motion.

    \item [2)] The generated video depth and optical flows are used to jointly control the video motion by regulating the object distances, spatial layouts and cross-frame correspondences. To preserve fine details and improve temporal consistency, different frames are aligned during both latent video generation and decoding stages.

    \item [3)] Experiments show that our method can generate videos with promising prompt consistency, frame consistency and visual quality, in both text-to-video and image-to-video generation in the open domain.
\end{itemize}

%% file: sec_eccv_cameraready/2_relatedwork.tex
\section{Related Work}
\label{sec:relatedwork}

\paragraph{Depth and optical flow generation.} As two of the fundamental computer vision problems, depth estimation and optical flow estimation have become hot topics for many years~\cite{ranftl2020towards, li2023learning, dosovitskiy2015flownet} and are widely used in video tasks such as video super-resolution~\cite{cao2021video,cao2022towards}, deblurring~\cite{liang2022vrt,liang2022rvrt}, denoising~\cite{cao2022learning} and frame interpolation~\cite{bao2019depth}. However, to the best of our knowledge, there are nearly no attempts in generating depth maps or optical flows for videos given a conditional input such as text or image. One related method~\cite{chen2023motion} proposes to generate the dense flow map from the sparse flow map input, but it needs strokes by human.

\paragraph{Video generation.}
Video generation aims to generate videos, mostly under guidance such as text~\cite{ho2022video, ho2022imagen}. Although GAN-based models~\cite{goodfellow2020generative, vondrick2016generating, saito2017temporal, tulyakov2018mocogan,clark2019adversarial} achieved good results in the past years, most recent video generation models are based either on sequence-to-sequence models~\cite{chang2022maskgit, chang2023muse} or on diffusion models~\cite{sohl2015deep,song2020denoising,song2020score,rombach2022high,zhang2023text2nerf,bar2024lumiere,zhang2024videoelevator}. The former line of work first tokenize each video frame into a sequence of discrete tokens and then transform the video generation problem to a sequence-to-sequence translation problem. It could be further divided as autoregressive~\cite{yan2021videogpt,wu2022nuwa,ge2022long,hong2022cogvideo} and non-autoregressive models~\cite{yu2023magvit,villegas2022phenaki}.

The other line of work use the diffusion process for video generation. Ho~\etal~\cite{ho2022video} propose a spatial-temporal factorized 3D UNet by adding temporal blocks for video generation, as a natural extension of the standard image diffusion model~\cite{sohl2015deep}. Similarly, Ho~\etal~\cite{ho2022imagen} apply the same idea to the cascaded image diffusion model Imagen~\cite{saharia2022photorealistic}, while He~\etal~\cite{he2022latent}, Zhou~\etal~\cite{zhou2022magicvideo} and Wang~\etal~\cite{wang2023modelscope} apply it to the latent space~\cite{rombach2022high}. Different from above methods that use spatial-temporal factorized 3D UNet, An~\etal~\cite{an2023latent} keep using 2D UNet and enable motion learning by shifting the feature channels along the temporal dimension. Some other methods take the redundancy of videos into consideration. Luo~\etal~\cite{luo2023videofusion} represent each frame as the addition of the base frame and residue, while Ge~\etal~\cite{ge2023preserve} encode each frame as the concatenation of the shared latent variable and an individual latent variable. 

In particular, training video generation models requires large-scale annotated video data, which are often not publicly available~\cite{ho2022video,ho2022imagen,villegas2022phenaki,zhou2022magicvideo}. Therefore, some methods instead propose to generate a video from one of its frames~\cite{singer2022make, esser2023structure}, as it is fully self-supervised and unlabelled videos are widely available. Singer~\etal~\cite{singer2022make} propose a cascaded architecture to generate high-spatiotemporal-resolution videos given an image embedding and a desired frame rate. Esser~\cite{esser2023structure} propose a video editing framework based on the edited image embedding and the depth of the original video. Some other methods try to generate or edit videos based on pre-trained image diffusion models~\cite{guo2023animatediff,zhang2023controlvideo,qi2023fatezero,ceylan2023pix2video,liu2023video,khachatryan2023text2video,geyer2023tokenflow,wu2023tune}. They often freeze or only optimize some of the model parameters on a single video input. Since our method is proposed for open-domain video generation, the comparison with these zero-shot, one-shot or few-shot methods are omitted here.

\paragraph{Perceptual Video Compression.}
Generating videos in the latent space is either necessary for sequence-to-sequence models or preferred in diffusion models for the sake of computation burden~\cite{rombach2022high}. Many existing models~\cite{zhou2022magicvideo,yu2023video,he2022latent,li2023videogen} directly use the pre-trained image VQVAE~\cite{van2017neural} or VQGAN~\cite{esser2021taming} for encoding and decoding videos, where each frame of the video is processed independently. To improve reconstruction quality, Yu~\etal~\cite{yu2023magvit}, He~\etal~\cite{he2022latent} and Ruben~\etal~\cite{villegas2022phenaki} propose several 3D autoencoders to compress the video both spatially and temporally, while Blattmann~\etal~\cite{blattmann2023align} fix the encoder of VQGAN and only add additional temporal layers in the decoder.

%% file: sec_eccv_cameraready/3_method.tex
\section{Method}
Due to the lack of large-scale high-quality paired text-video datasets, we limit our setting to uncaptioned video data and try to generate a video from one of its key frames in a fully self-supervised way. To achieve this, we propose a \textbf{Mo}tion-aware \textbf{Video} generation framework (referred to as MoVideo) with explicit motion modelling. It consists of four stages: key frame generation, video depth and optical flow generation, depth and optical flow-based video generation, and optical flow-augmented video decoding, as shown in Fig.~\ref{fig:teaser}. Any public available diffusion model could be used to generate the key frame based on the text prompt or we directly use an existing image as the key frame. The next three stages, including a brief introduction to the diffusion models, are detailed as below.

\subsection{Preliminaries on Diffusion Models}
Diffusion models~\cite{sohl2015deep,ho2020denoising,song2020denoising,song2020score} are probabilistic models that learn a data distribution $p(x)$ by gradually denoising a normally distributed variable $x_T\sim\mathcal{N}(0,1)$ to obtain the original data $x_0\sim p(x)$. In the forward diffusion process, we define a fixed Markov chain of length $T$ as
\begin{align}
q(x_t|x_{t-1}) = \mathcal{N}(x_t; \sqrt{\alpha_t} x_{t-1}, (1-\alpha_t)I),
\end{align}
where each $x_t$ is obtained by adding a small normally distributed noise to $x_{t-1}$. $\alpha_1$ is set to be slightly smaller than 1 and $\alpha_t$ is scheduled to decrease gradually for $t=1,...,T$. When $T$ is large enough, \eg, $T=1000$, $x_T$ is nearly independent of $x_0$ and meets $x_T\sim\mathcal{N}(0,1)$.

Since estimating the conditional probability $q(x_{t-1}|x_t)$ is intractable, we approximate it by learning the distribution $p_\theta(x_{t-1}|x_t)$ with parameters $\theta$. Starting from a random noise $x_T\sim\mathcal{N}(0,1)$, we gradually remove noises with a Markov chain in $T$ steps based on the learned Gaussian transitions. The reverse denoising process is defining as
\begin{align}
{p_\theta}(x_{t-1} | x_t) &= \mathcal{N}\big(x_{t-1};
{\mu_\theta}(x_t, t), {\Sigma_\theta}(x_t, t)\big) \\
{p_\theta}(x_{0:T}) &= {p_\theta}(x_T) \prod_{t = 1}^{T} {p_\theta}(x_{t-1} | x_t) \\
{p_\theta}(x_0) &= \int{p_\theta}(x_{0:T}) dx_{1:T}~\label{eq:p_x_0}
\end{align}

To optimize the negative log-likelihood of~\eqref{eq:p_x_0}, we can derive its variational lower bound by the reparameterization trick~\cite{kingma2013auto,sohl2015deep}. In practice, we empirically use a simplified reweighted variant of variational lower bound~\cite{ho2020denoising} as
\begin{align}
L
= \mathbb{E}_{{x}_0, t \sim [1, T], {\epsilon\sim\mathcal{N}(0,1)}} \Big[\|{\epsilon} - {\epsilon}_\theta({x}_t, t)\|^2_2 \Big],
\end{align}
where ${\epsilon}_\theta(\cdot)$ is parameterized by a denoising UNet~\cite{ronneberger2015u} that takes the noisy sample ${x}_t=\sqrt{\bar{\alpha}_t}{x}_0 + \sqrt{1 - \bar{\alpha}_t}{\epsilon}$ and the corresponding diffusion timestep $t$ as inputs, and outputs the predicted noise. $\bar{\alpha}_t$ is defined as $\prod_{i=1}^t \alpha_i$.

To boost the efficiency, some methods propose to conduct the diffusion process in a compressed latent space defined by a pre-trained autoencoder~\cite{esser2021taming}. When we apply it to the video domain, we can use the image encoder $\mathcal{E}$ to encode a video $x\in\mathcal{R}^{F\times H\times W\times C}$ frame by frame to a latent variable $z=\mathcal{E}(x)\in\mathcal{R}^{F\times h\times w\times c}$, and use the decoder $\mathcal{D}$ to reconstruct the video $\mathcal{D}(z)\approx x$. $F$, $H$, $W$ and $C$ are video frame number, height, width and channel number, respectively, in the pixel space, while $h$, $w$ and $c$ are latent variable height, width and channel number, respectively, in the latent space. In particular, one can further speed up the reverse denoising process by the deterministic sampling~\cite{song2020denoising}.

\begin{figure}
\begin{minipage}[c]{0.49\textwidth}
\begin{overpic}[width=6.cm]{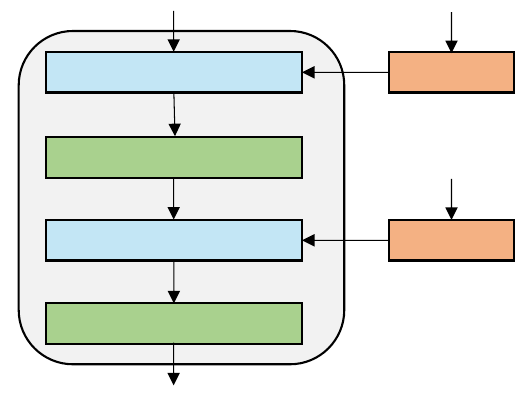}
\put(31,-3){\color{black}{\tiny{${x}$}}}
\put(31,75){\color{black}{\tiny{${x}$}}}
\put(80.5,75){\color{black}{\tiny{${fps}$}}}
\put(80.5,43){\color{black}{\tiny{$x_{key}$}}}
\put(13.7,60){\color{black}{\tiny{2D spatial convolution}}}
\put(12,44.1){\color{black}{\tiny{1D temporal convolution}}}
\put(15.5,28.5){\color{black}{\tiny{2D spatial attention}}}
\put(14,12.8){\color{black}{\tiny{1D temporal attention}}}
\put(80.5,60){\color{black}{\tiny{MLP}}}
\put(78.5,28.5){\color{black}{\tiny{Encoder}}}
\end{overpic}
  \end{minipage}\hfill
  \begin{minipage}[c]{0.49\textwidth}
    \caption{The basic spatio-temporal block for building the 3D denoising UNet. We add temporal modules, including temporal convolution and temporal attention layers after spatial convolution and spacial attention layers. The ${fps}$ is encoded by a multi-layer perceptron and then added to the feature after 2D convolution, while key frame $x_{key}$ is encoded by the image encoder from an open-sourced pretrained image-text bi-encoder model and then injected to the 2D spatial attention layer by cross attention.} \label{fig:3dunet}
  \end{minipage}
\end{figure}

\subsection{Video Depth and Optical Flow Generation}
\label{sec:depth_generation}
Compared with image generation, the main challenge in video generation lies in how to generate motions with good temporal consistency. To describe motions in a video, the most widely used way is calculating the optical flows that represent the correspondences in a video. However, optical flows cannot handle occlusions well and often lead to blurry or deformed motion boundaries ~\cite{bao2019depth}. To remedy this, we propose to combine the optical flows with video depth maps that reflect the object distances and spatial layout of each frame, providing accurate information for boundary movements in a video. It is noteworthy that merely using depth maps might be insufficient as they cannot guarantee the temporal consistency of fine details such as textures and colors in different frames. There, before video generation, we jointly generate the video depth and optical flows based on the key frame in this subsection.

Formally, given a center key frame $x_{key}\in\mathcal{R}^{H\times W\times C}$ from a video $x$, we first use an open-sourced pretrained image-text bi-encoder model $\mathcal{C}$ to extract the image embedding $\mathcal{C}(x_{key})$ before the last average pooling layer. Then, $\mathcal{C}(x_{key})$ is used as the conditional input to control the contents of the generated video depth and optical flows, and the frames per second ($fps$) is used as an additional condition for controlling the motion magnitude. With these two conditions, we design a diffusion model to learn a joint distribution of depth and optical flows as
\begin{align}
    d, o^{i2v},o^{v2i}\sim p_\theta\big(d, o^{i2v},o^{v2i}|\mathcal{C}(x_{key}),fps\big),
\end{align}
where $d\in\mathcal{R}^{T\times H\times W\times 1}$ is the video depth that includes the depth maps for each frame, $o^{i2v}\in\mathcal{R}^{T\times H\times W\times 2}$ is the image-to-video optical flow from the key frame to all frames in the video, and $o^{v2i}\in\mathcal{R}^{T\times H\times W\times 2}$ is the video-to-image optical flow. In detail, for the architecture of the denoising UNet, the original 2D network is upgraded to a 3D network that is able to predict the video noise added to the concatenation of video depth and optical flow tensors. As shown in Fig.~\ref{fig:3dunet}, after each spatial block (including 2D spatial convolution layers and 2D spatial attention layers), we add temporal blocks that consist of several 1D temporal convolution layers and 1D temporal attention layers. The alternative stacking of spatial and temporal blocks allows the network to capture spatio-temporal correlations in videos and generate temporally consistent motions. As for the conditioning mechanism, the image embedding is injected to the model by cross attention~\cite{vaswani2017transformer}, after self attention in the 2D spatial attention block. The $fps$ is first transformed to a deep feature by a multi-layer perceptron (MLP) and then added to network after 2D convolution layers.

\paragraph{Remarks.} Due to the characteristic difference from RGB video data, there are some remarks on designing a diffusion model for video depth and optical flow generation as below.

\begin{figure*}[t]
\centering
\includegraphics[width=12cm]{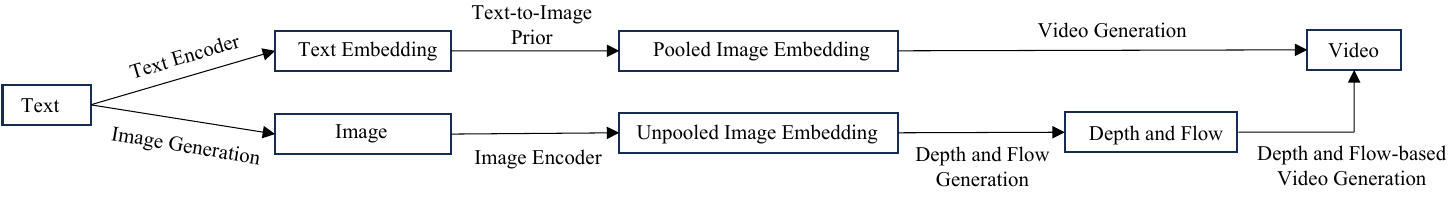}
\captionsetup{type=figure}
\caption{The comparison on different architectures for text-to-video generation without text-video training pairs. As the top route shows, some methods~\cite{singer2022make,esser2023structure} first encode the text with the text encoder from an open-sourced pretrained image-text bi-encoder model and then use a text-to-image prior~\cite{ramesh2021zero,ramesh2022hierarchical} to transform it to the pooled image embedding, which is used as the condition to guide the generation of video. Instead, we propose to first generate an image by a public text-to-image latent diffusion model and extract its unpooled image embedding that preserves spatial layout and local details of the image, based on which we generate the depth and optical flow of the video and then use them to guide video generation.}
\label{fig:archi_compare}
\end{figure*}

\hspace{0.5em}1) \emph{Use unpooled image embedding.} Generally, we use outputs of the last average pooling layer of the image encoder as image embeddings~\cite{singer2022make,esser2023structure}, as they are aligned with text embeddings and could be obtained by inputting the text embedding to the text-to-image prior~\cite{ramesh2022hierarchical} during testing, as shown in the top of Fig.~\ref{fig:archi_compare}. However, image semantics, spatial layout and local details might be lost after pooling, posing challenges in generating depth and optical flows. Hence, we use the unpooled image embedding. 
Additionally, sinusoidal positional encoding~\cite{vaswani2017transformer} is added to the image embedding for encoding spatial information.
Nevertheless, above design brings another challenge on training a text-to-image prior for sampling unpooled image embeddings. To tackle with it, as shown in the bottom of Fig.~\ref{fig:archi_compare}, we propose to first generate a RGB image from the text input by a public text-to-image latent diffusion model and then extract its image embedding, allowing us to utilize the strong image diffusion prior to generate the basic semantics of video.

\hspace{0.5em}2) \textit{Use image-to-video and video-to-image optical flows.} There are two  common choices of describing the video motion with optical flows: flows between neighboring frame pairs or between the key frame and other frames. We use the second design due to following two reasons. First, we can directly obtain all frames from the key frame by optical flow-based warping, avoiding the accumulation of warping errors during propagation. Second, since optical flow only reflects relative movements, our design makes it possible to normalize all flows together, which is found to be critical in flow generation.

\hspace{0.5em}3) \textit{Use normalized depth and optical flows.}
Since depth maps and optical flows have different numeric ranges, we normalize them separately to be within -1 and 1 as
\begin{align}
\widetilde{d}&=2\times\frac{d-min(d)}{max(d)-min(d)}-1,\\
o_{max}&=max\big(max(\|o^{i2v}\|_2),max(\|o^{v2i}\|_2)\big),\\
\widetilde{o^{i2v}}, \widetilde{o^{v2i}}&=\frac{o^{i2v}}{o_{max}},\frac{o^{v2i}}{o_{max}},
\end{align}
where $\|\cdot\|_2$ denotes the norm of the optical flow motion vector. However, after normalization, we cannot obtain the original optical flow values and use them for warping, as maximum flow $o_{max}$ is unknown in inference. To remedy this, based on the consistency of video depth maps, we propose an optimization-based method to infer $o_{max}$ from normalized depth maps and optical flows as
\begin{equation}
\begin{aligned}
o_{max}=\argmin_{o_{max}} \sum_{f=1}^F \{\|\widetilde{d}_f - \mathcal{W}(\widetilde{d}_{key}, \widetilde{o^{i2v}_f}*o_{max})\|_2+\\
\|\widetilde{d}_{key} - \mathcal{W}(\widetilde{d}_{f}, \widetilde{o^{v2i}_f}*o_{max})\|_2\},
\end{aligned}
\end{equation}
where $\mathcal{W}$ is the flow-based warping operation. $o_{max}$ is optimized by minimizing the above cost function.

\subsection{Depth and Flow-based Video Generation}
\label{sec:video_generation}
Given the key frame $x_{key}$, the frames per second $fps$, video depth $d$, image-to-video optical flow $o^{i2v}$ and video-to-image optical flow $o^{v2i}$ generated in subsection~\ref{sec:depth_generation}, we first compress the key frame to the latent space as $z_{key}=\mathcal{E}(x_{key})$ with the pre-trained latent encoder~\cite{esser2021taming}, and then obtain the occlusion mask $m$ and the warped latent video $\widetilde{z}$ as
\begin{align}
m_f&=\{\|o^{i2v}_f + \mathcal{W}(o^{v2i}_f, o^{i2v}_f)\|_2<\delta_f\},\\
\widetilde{z_f}&=\mathcal{W}(z_{key}, {o^{i2v}_f}) * m_f,
\end{align}
where $\delta_f=\alpha(\|o^{i2v}_f\|_2+\|o^{v2i}_f\|_2)+\beta$ is the threshold for non-occluded regions~\cite{meister2018unflow}. $\alpha$ and $\beta$ are set as 0.01 and 0.5, respectively. $f=1,...,F$ represents frame index.

Next, we turn to diffusion models again to learn a conditional distribution of the latent video $z$ as
\begin{align}
    z\sim p_\theta\big(z|\mathcal{C}(x_{key}),fps,d, \widetilde{z},m\big),
\end{align}
where $\mathcal{C}(x_{key})$ is still the image embedding of the key frame as in subsection~\ref{sec:depth_generation}. Similarly, we upgrade the 2D UNet to a 3D UNet by inserting temporal layers as shown in Fig.~\ref{fig:3dunet}. The parameters of the spatial layers are initialized with the original 2D UNet from an open-sourced pretrained latent diffusion model and optimized with one tenth of the global learning rate to preserve the learned image diffusion prior, while the newly added layers are initialized as identical mappings. The conditioning mechanisms for $fps$ and $\mathcal{C}(x_{key})$ are kept the same as in~\ref{sec:depth_generation}. Besides, we concatenate $d$, $\widetilde{z}$ and $m$ with $z$ as the input to the UNet to guide the video generation process for different purposes. The video depth $d$ controls the video motion and the rough layout of each frame. The warped latent video $\widetilde{z}$ provides semantic information and local details, and it also helps to improve the temporal consistency, especially for texture and color consistencies based on the correspondence information from the optical flow. The occlusion mask $m$ is used to indicate occluded regions, which can tell the network whether the information from $\widetilde{z}$ could be trusted or not.

Particularly, when we concatenate the warped latent video $\widetilde{z}$ as the input condition, we empirically found that it leads to unsatisfactory motions in the generated video, possibly because the reconstruction process is sometimes misled by the imperfect motions from $\widetilde{z}$. To alleviate the problem, we randomly mask $\widetilde{z}$ as all-zeros with a probability of 0.5, which prevents the network from learning bad motions of $\widetilde{z}$ but still allows it to benefit from the semantic information and local details of $\widetilde{z}$.

\subsection{Optical Flow-Augmented Video Decoding}
\label{sec:decoding}
\begin{figure}[!t]
  \begin{minipage}[c]{0.55\textwidth}
    \begin{overpic}[width=7.cm]{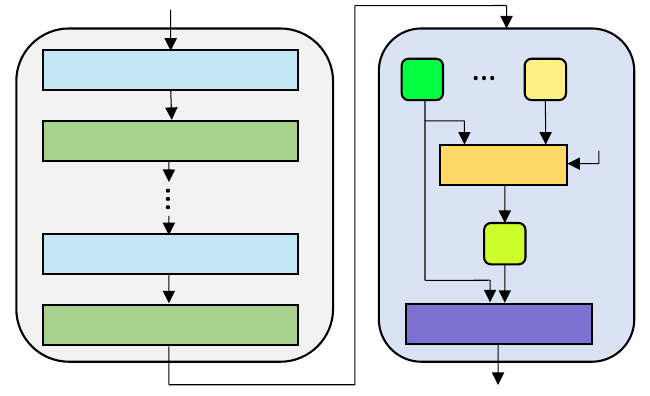}
\put(82,22.8){\color{black}{\scriptsize{$\widetilde{z_f}$}}}
\put(88.6,40.7){\color{black}{\scriptsize{${o^{i2v}_f}$}}}
\put(63,53.8){\color{black}{\scriptsize{${z_f}$}}}
\put(80.3,54){\color{black}{\scriptsize{$z_{key}$}}}
\put(9.5,49.5){\color{black}{\tiny{2D spatial convolution}}}
\put(7.9,38.1){\color{black}{\tiny{1D temporal convolution}}}
\put(9.5,21){\color{black}{\tiny{2D spatial convolution}}}
\put(7.9,10.1){\color{black}{\tiny{1D temporal convolution}}}
\put(70.,34.8){\color{black}{\tiny{alignment}}}
\put(63.2,10.3){\color{black}{\tiny{feature refinement}}}
\end{overpic}
  \end{minipage}\hfill
  \begin{minipage}[c]{0.38\textwidth}
    \caption{The basic block for building the optical flow-augmented video decoding model. We add temporal convolution layers after spatial convolution layers to extract spatio-temporal video features. After that, with the optical flow ${o^{i2v}_f}$, we align the key frame $z_{key}$ towards the each frame as $\widetilde{z_f}$, which is concatenated with $z_f$ for feature refinement.}
\label{fig:3d_decoding}
  \end{minipage}
\end{figure}

After obtaining the latent video $z$ from the last subsection~\ref{sec:video_generation}, the final step is to decode it to the pixel space as video $x$. Since different video frames are highly related but misaligned, we propose to align different frames for joint decoding. As shown in Fig.~\ref{fig:3d_decoding}, we upgrade the 2D decoder in~\cite{esser2021taming} to a 3D decoder by inserting temporal convolution layers after spatial convolution layers. Then, we fuse cross-frame information by explicitly aligning the key frame feature $z_{key}$ towards the each frame feature $z_f$ with flow-guided deformable convolution~\cite{chan2021basicvsr++,dai2017deformable} as
\begin{align}
    {z'_f}&=\mathcal{W}(z_{ref},o^{i2v}_f),\\
    s_{f},a_{f}&={CNN}(Concat(o^{i2v}_{f},{z_f}', z_f )),\\
    \widehat{z_f}&={DC}(z_{ref},o^{i2v}_f+s_{f},a_f),
\end{align}
where the offset $s_f$ and mask $a_{f}$ are estimated according the concatenation of optical flow $o^{i2v}_{f}$, the warped feature ${z_f}'$ and the frame $z_f$. ${CNN}$ is a neural network with several $3\times 3$ convolution layers and $DC$ is the deformable convolution. Next, we concatenate the aligned feature $\widehat{z_f}$ with $z_f$ and use several ResNet blocks~\cite{he2016resnet} (denoted as $\mathcal{R}$) to refine the feature as
\begin{align}
    z_f = z_f + \mathcal{R}(Concat(\widehat{z_f}, z_f)),
\end{align}
which is used for each stage of the decoder. Hence, the decoding of every frame $f$ is aware of other frames and especially obtains information from the key frame.

In training, we use a combination of pixel loss, perceptual loss~\cite{johnson2016perceptual} and GAN loss~\cite{goodfellow2020generative} following VQGAN~\cite{esser2021taming}. To improve temporal coherence, we further extract features from S3D~\cite{xie2018rethinking} for 3D perceptual loss and upgrade the 2D discriminator to a 3D discriminator for the 3D GAN loss by replacing the 2D convolutions with 3D convolutions. We train the decoder with a combination of losses between the decoded video $\hat{x}$ and ground-truth video $x$ as:
\begin{equation}
  \begin{aligned}
    L = {L_1}(\hat{x},x)+&\lambda_1\sum_{f=1}^FL_{percep\_2d}(\hat{x},x)+\lambda_2 L_{percep\_3d}(\hat{x},x)\\
    +&\lambda_3\sum_{f=1}^FL_{GAN\_2d}(\hat{x},x)+\lambda_4 L_{GAN\_3d}(\hat{x},x).
\end{aligned}  
\end{equation}

%% file: sec_eccv_cameraready/4_experiment.tex
\section{Experiments}
\subsection{Experimental Setup}
\paragraph{Architecture.}
For video generation, we use the spatial blocks from the 2D UNet of an open-sourced pretrained latent diffusion model and add the same numbers of temporal blocks with similar channel numbers to upgrade it to a 3D UNet. For video depth and optical flow generation, we reduce the channel numbers by half as depth maps and optical flows have less details. For video decoding, we upgrade the 2D VQGAN decoder~\cite{esser2021taming} to a 3D decoder in a similar way. The details on the networks are provided in the supplementary.

\paragraph{Training and testing.}
In training, we use the similar training set as in~\cite{singer2022make} and randomly cut 17-frame video clips with a random $fps$ between 3 and 30. Each video clip is resized to ensure its shorter side measures 256 pixels, followed by a center-crop of size $256\times 256$. The depth maps and optical flows are estimated by open-sourced pretrained depth and optical flow models. The depth and optical flow generation model, video generation model (includes the fine-tuning stage) and video decoding model are trained separately with different training iterations, batch sizes and optimizer settings. In testing, we generate $17\times 256\times 256$ videos at a $fps$ of 3. More details are summarized in the supplementary.

\begin{figure}[!t]
  \begin{minipage}[c]{0.69\textwidth}
    \begin{overpic}[width=8cm]{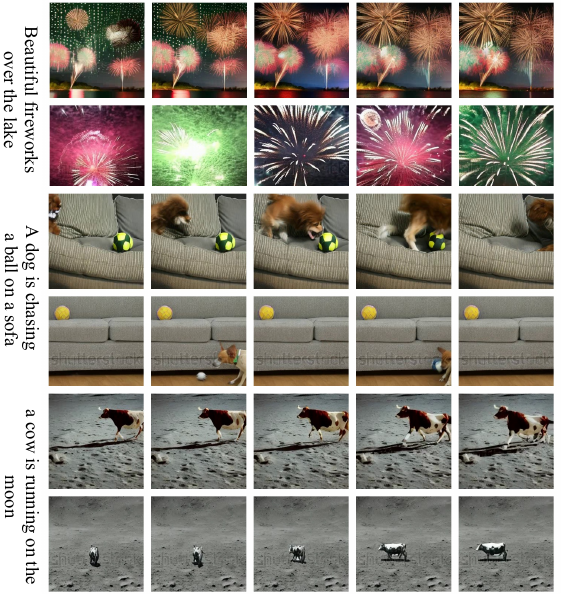}
\end{overpic}
  \end{minipage}\hfill
  \begin{minipage}[c]{0.29\textwidth}
    \caption{Visual comparison on text-to-video generation. For each example, the first row is from our method, while the second row is from VideoDiffusion~\cite{yu2023video}. {More visual comparisons, including video results, are provided in the supplementary.}}
\label{fig:visual_t2v_second}
  \end{minipage}
\end{figure}

\subsection{Text-to-Video Generation}
We test the text-to-video generation ability of our model on multiple text prompts in Figures~\ref{fig:visual_t2v_second}. We mainly compare our method with one of the recent best models VideoFusion~\cite{luo2023videofusion}, as it is the only open-sourced text-to-video model during the time when we did experiments (to the best of our knowledge) and it uses the same publicly available training set as in our model. As we can see, our model can generate high-quality visually-pleasing videos with good consistency in three aspects. First, it is consistent with the semantics of text prompts, including the objects (\eg, ``fireworks'' and ``lake'' in the first example)  and the motion (\eg, ``chasing'' in the second example). In contrast, VideoFusion fails to generate the ``lake'' as the background or show the ``chasing'' action between the ``dog'' and ``ball''. Second, across different frames, the generated objects of our method are consistent (\eg, the same ``fireworks'' in the first example), while VideoFusion generates ``fireworks'' with different colors and shapes for neighbouring frames. Third, our method is able to generate natural consistent motions without sudden large movements, \eg, the ``chasing'' of ``dog'' is natural), but the ``dog'' generated by VideoFusion may disappear suddenly in some frames. The visualization of intermediate results and more comparisons are provided in the supplementary due to page limit.

For quantitative comparison, we compare our method on zero-shot text-to-video generation on UCF-101~\cite{soomro2012ucf101} and MSR-VTT~\cite{xu2016msr}. As shown in Tables~\ref{tab:quan_t2v_ucf101} and~\ref{tab:quan_t2v_msrvtt}, our method achieves best performance on all metrics. Besides, we choose 100 text prompts to generate videos in the open domain. As shown in Table~\ref{tab:quan_t2v}, generated videos of our method is more consistent with the prompts and across the frames. It is preferred by 81.3\% of the 30 users on average during user study.

\begin{table}[!t]\small
\captionsetup{font=small}
\begin{minipage}[c]{0.44\textwidth}
\caption{Quantitative comparison of zero-shot text-to-video generation on UCF-101~\cite{soomro2012ucf101}.}
\label{tab:quan_t2v_ucf101}
\begin{center}
\begin{tabular}{|c|c|c|}
\hline
Method & IS$\uparrow$ & FVD$\downarrow$\\
\hline\hline
CogVideo~\cite{hong2022cogvideo}  & 25.27 & 701.59 \\
Make-A-Video~\cite{singer2022make}  & 33.00 & 367.23 \\
Video LDM~\cite{blattmann2023align}  & 33.45 & 550.61 \\
MoVideo (ours) & 34.13 & 313.41 \\
\hline
\end{tabular}
\end{center}
  \end{minipage}\hfill
\begin{minipage}[c]{0.54\textwidth}

\caption{Quantitative comparison of zero-shot text-to-video generation on MSR-VTT~\cite{xu2016msr}.}
\label{tab:quan_t2v_msrvtt}
\begin{center}
\begin{tabular}{|c|c|c|}
\hline
Method & FID$\downarrow$ & CLIPSIM$\uparrow$\\
\hline\hline
N{\"U}WA~\cite{wu2022nuwa} & 47.68 & 0.2439\\
CogVideo~\cite{hong2022cogvideo} & 23.59 & 0.2631 \\
Latent-Shift~\cite{an2023latent} & 15.23 & 0.2773 \\
Make-A-Video~\cite{singer2022make} & 13.17 & 0.3049 \\
Video LDM~\cite{blattmann2023align}  & - & 0.2929 \\
MoVideo (ours) & 12.71 & 0.3213 \\
\hline
\end{tabular}
\end{center}
\end{minipage}
\end{table}

\begin{table}[!t]\small
\captionsetup{font=small}
\caption{Quantitative comparison of open-domain text-to-video generation.}
\label{tab:quan_t2v}
\begin{center}
\begin{tabular}{|c|c|c|c|}
\hline
Method & \makecell{Frame Consistency$\uparrow$} & \makecell{Prompt Consistency$\uparrow$} & \makecell{Preference Rate$\uparrow$} \\
\hline\hline
\scalebox{1}{VideoDiffusion~\cite{luo2023videofusion}}  & 0.9759 & 0.3143 & 18.7\%\\
MoVideo (ours) & 0.9867 & 0.3631 & 81.3\%\\
\hline
\end{tabular}
\end{center}
\end{table}

\begin{table}[!t]\small
\captionsetup{font=small}
\caption{Quantitative comparison of image-to-video generation on DAVIS~\cite{khoreva2018davis}.}
\label{tab:quan_i2v}
\begin{center}
\begin{tabular}{|c|c|c|c|c|}
\hline
Method & PSNR$\uparrow$ & LPIPS$\downarrow$ & FID$\downarrow$ & FVD$\downarrow$  \\
\hline\hline
{Gen1~\cite{esser2023structure}}  & 28.39  & 0.290 & 149.73 & 973.71\\ %
MoVideo (ours) & 29.88  & 0.079 & 40.14 & 335.53 \\ %
\hline
\end{tabular}
\end{center}
\end{table}

\subsection{Image-to-Video Generation}
\begin{figure}[!t]
  \begin{minipage}[c]{0.69\textwidth}
\begin{overpic}[width=8cm]{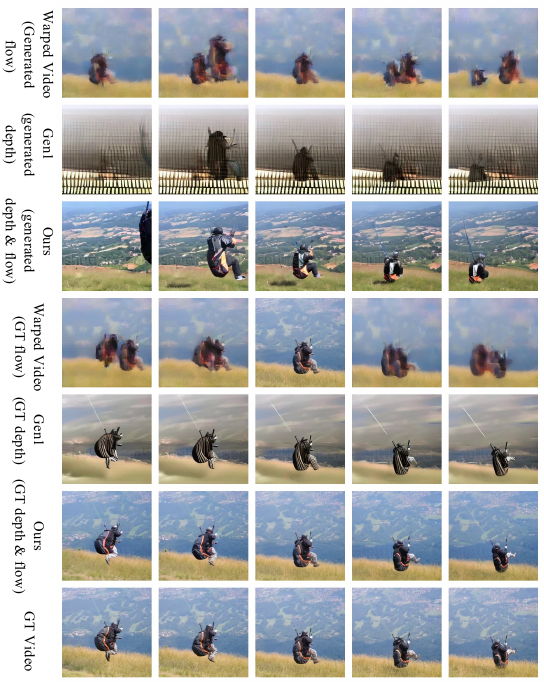}
\end{overpic}
  \end{minipage}\hfill
  \begin{minipage}[c]{0.29\textwidth}
\caption{Visual comparison on image-to-video generation. The first three rows are guided by the generated depth and optical flows, while the rest rows are guided by the ground-truth (GT) ones. {More visual comparisons, including video results, are provided in the supplementary.}}
\label{fig:visual_i2v}
  \end{minipage}
\end{figure}

We show the results for image-to-video generation in Fig.~\ref{fig:visual_i2v} on the DAVIS~\cite{khoreva2018davis} dataset. We use the flow-based warping as the baseline and mainly compare our method with the recent best method Gen1~\cite{esser2023structure}, which is re-implemented by us and trained with the same training settings for fair comparison. As shown in the figure, the generated video of our method is almost the same as the ground-truth video, while the Gen1 changes the background and object appearance. The quantitative results in Table~\ref{tab:quan_i2v} further validate our observation from the aspects of fidelity (reflected by PSNR) and perceptual quality (reflected by LPIPS~\cite{zhang2018lpips}, FID~\cite{heusel2017gans} and FVD~\cite{unterthiner2018towards}). When we use the generated depth and optical flows from the image, our method is able to generate a video with similar semantics but different motions.

\subsection{Ablation Studies}
\hspace{1.2em} \emph{Ablation study on the warped video.}
We obtain a warped video by warping the key frame with the image-to-video optical flow, and then use it to guide the video generation. As shown in Table~\ref{tab:ablation_warped_occlusion}, there exists a large margin between with and without warped video, which indicates that the concatenation of the warped video plays a critical role in the model performance. In fact, without warped video, the model cannot preserve the textures and colors. A visual example is provided in the supplementary.

\emph{Ablation study on the occlusion mask.}
The occlusion mask is concatenated as the conditional input to help the model identify the occluded regions. Without the occlusion mask, the performance drops as show in Table~\ref{tab:ablation_warped_occlusion}, possibly due to the wrong correspondences appearing in the occluded regions of videos. In this case, we found that the ghosting artifact often appears, as provided in the supplementary.

\emph{Ablation study on flow-augmented video decoding.}
The optical flow is used to align different frames during decoding. To validate its effectiveness, we encode the videos from DAVIS~\cite{khoreva2018davis} with the pre-trained image encoder~\cite{esser2021taming}, and use different decoders to decode the video. As shown in Table~\ref{tab:ablation_decoder}, when upgrading the 2D decoder to a 3D one, the perceptual metrics become better with slight drop in PSNR. When using the proposed flow-augmented video decoder, we achieve the best results on all perceptual metrics, indicating its ability to decode better visually-pleasing videos.

To to page limit, the accompanying qualitative results, video results, as well as more ablation studies are provided in the supplementary.

\begin{table}[!t]\small
\captionsetup{font=small}
\caption{Ablation study on video decoding.}
\label{tab:ablation_decoder}
\begin{center}
\begin{tabular}{|c|c|c|c|}
\hline
Decoder & 2D decoder & 3D decoder & MoVideo (ours) \\
\hline\hline
PSNR$\uparrow$ & 30.11 & 29.99 & 29.88\\
LPIPS$\downarrow$ & 0.093 & 0.087 & 0.079\\
FID$\downarrow$ & 43.28 & 42.71 & 40.14\\
FVD$\downarrow$ & 401.66 & 379.73 & 335.53 \\
\hline
\end{tabular}
\end{center}
\end{table}

\begin{table*}[!t]\small
\captionsetup{font=small}
\caption{Ablation study on warped video and occlusion mask.}
\label{tab:ablation_warped_occlusion}
\begin{center}
\begin{tabular}{|c|c|c|c|c|c|}
\hline
Warped Video & Occlusion Mask & PSNR$\uparrow$ & LPIPS$\downarrow$ & FID$\downarrow$ & FVD$\downarrow$  \\
\hline
  & \checkmark &  27.92 &  0.303  &  153.73 & 796.82 \\ 
\checkmark  &  & 29.45 &  0.093  & 45.61  & 343.90  \\
\checkmark & \checkmark & 29.88  & 0.079 & 40.14 & 335.53 \\
\hline
\end{tabular}
\end{center}
\end{table*}

%% file: sec_eccv_cameraready/5_conclusion.tex
\section{Conclusion}
In this paper, we proposed a motion-aware video generation framework (MoVideo) that consists of four stages: key frame generation, video depth and optical flow generation, depth and optical flow-based video generation, and optical flow-augmented video decoding. Based on an text-to-image generation model, we use the text prompt to generate an image as the key frame, which is used to guide the generation of video motion represented by video depth, image-to-video optical flow and video-to-image optical flow. These motion representations could be further utilized to guide the latent video generation and assist the video decoding processes. Experiments demonstrate that our method achieves state-of-the-art performance on both text-to-video and image-to-video generation.

\emph{Limitation and  potential negative impact}
Although our multi-stage design allows explicit motion modelling, control and guidance, it suffers from multiple training and inference steps. Besides, the performance is highly related to optical flow and depth estimation methods as we use predicted pseudo-labels in training. These limitations could be our future working directions. For potential negative impact, this model might have the risks of data breaches or misinformation.